\documentclass[letterpaper, 10 pt, conference]{ieeeconf}  

\IEEEoverridecommandlockouts                              

\usepackage{graphicx}
\usepackage{subcaption}
\usepackage{float}
\usepackage{hyperref}

\title{\LARGE \bf
A reconfigurable robot workcell for quick set-up of assembly processes
}

\author{Timotej Ga\v{s}par, Miha Deni\v{s}a and Ale\v{s} Ude
\\
\textit{Jo\v{z}ef Stefan Institute}\\
\textit{Ljubljana, Slovenia}\\
{\tt\small \{timotej.gaspar, miha.denisa, ales.ude\}@ijs.si}
\thanks{This work has been funded by the Horizon 2020 ICT-FoF Innovation Action no 680431,  ReconCell (A Reconfigurable robot workCell for fast set-up of automated assembly processes in SMEs).\newline\newline
The accompanying video is available here:\newline \href{www.ijs.si/~tgaspar/iros/reconcell_fof.mp4}{www.ijs.si/~tgaspar/iros/reconcell\_fof.mp4}}
}

\begin{document}

\maketitle
\thispagestyle{empty}
\pagestyle{empty}

\begin{abstract}
High volume production has been a prerequisite in order to invest into automation of the manufacturing process for decades. The high cost of setup and the inflexibility of classical automation meant that low batch productions, often present in Small and Medium-sized Enterprises (SMEs), were dismissed as potential end user of automation technologies. In this extended abstract we present the results of the ReconCell project whose objective was to develop a new type of highly reconfigurable robot workcell for fast set-up of automated assembly processes in SMEs. The high degree of reconfigurability was achieved by the developed reconfigurable hardware and the complementary reconfigurable software, while fast set-up was achieved with technologies for fast robot programming.
\end{abstract}



\section{INTRODUCTION}

Despite the large increase of robots in manufacturing industry, their introduction in a production line still comes with high investment costs. While this does not represent an significant issue for big enterprises, some \textit{Smaller and Medium-sized Enterprises} (SME) in the manufacturing industry, cannot afford such an investment. These high costs usually stem from the price of the necessary hardware and the time spent for the integration of the robotic system into the production line. 

Another notable trend in today's manufacturing is the ever so rapid changes in market demands. To maintain competitiveness in the diverse and global market, SMEs must follow these trends of automation. In this regard, the Reconfigurable Manufacturing Systems (RMS) paradigm advocates for manufacturing systems that can rapidly and efficiently adjust production capacity and functionality to meet sudden changes in market demands \cite{koren_1999}. Even if these systems are -- compared to more dedicated systems -- associated with lower throughput and more complex design, they provide added value in manufacturing processes where changes in production happen relatively often \cite{zhang_analytical_2006}. As SMEs represent an important factor in terms of manufacturing and employment, they can be perceived as a target group for RMS \cite{european_commission_factories_2013}. However, to make RMS more viable for SMEs, it is important to tackle the issue of the high investment costs of applying such systems to the manufacturing processes \cite{dietz_programming_2012}.


\begin{figure}[t]
	\begin{center}
		\includegraphics[width=.95\linewidth]{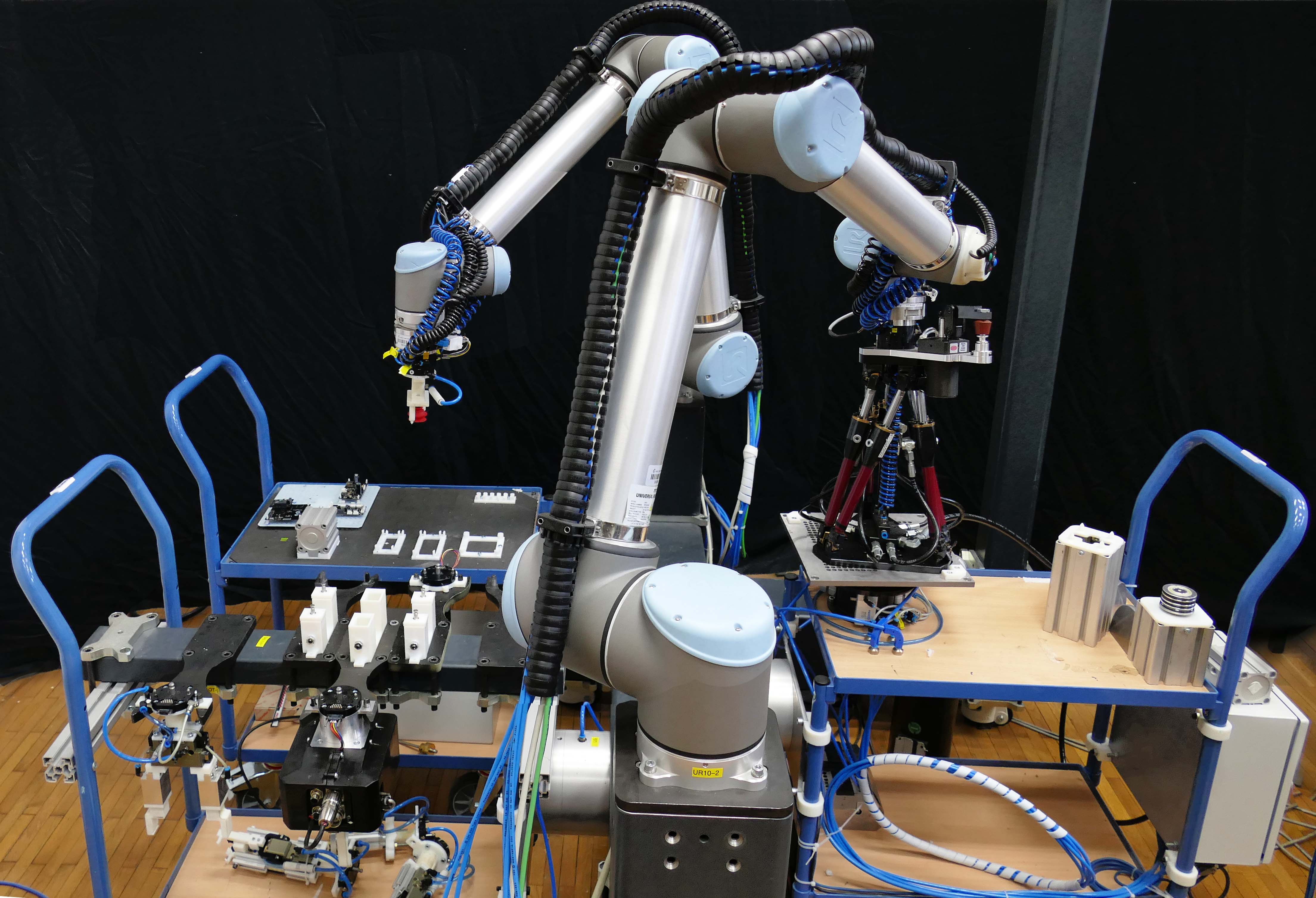}
		\caption{An overview of the reconfigurable robot workcell during an assembly process.}
		\label{fig:workcell-overview}
	\end{center}
\end{figure}

The objective of our work was to design and implement an RMS in shape of a reconfigurable robot workcell, suitable not only for large production lines but also for low-volume high-diversity production, which often takes place in Small and Medium Enterprises (SMEs). To achieve this, we developed novel hardware elements that offer cost efficient solutions to reconfigurability, complemented by a reconfigurable ROS-based software architecture. To increase the intuitiveness of our system and to accelerate setup time, we made use of the programming by demonstration methods to teach robots assembly skills. 

This extended abstract is organised as follows. Section \ref{sec:reconf-hardware} describes the novel approaches in hardware design for reconfigurable robot workcells. In Section \ref{sec:reconf-software} we present the underlying ROS-based software architecture of the cell. Technologies for fast set-up times and intuitive robot programming are presented in Section \ref{sec:task-programming}. The results the implementation of the proposed system and the concluding remarks are provided in Section \ref{sec:conclusion}

\section{RECONFIGURABLE HARDWARE}\label{sec:reconf-hardware}

When developing a reconfigurable robot workcell, it is necessary to consider the desired physical properties of the overall system, e.\,g. size, stiffness, robot workspace, etc. The available factory space plays a significant role in determining the layout of the workcell. It is equally important to ensure, that the workcell can be integrated into an existing manufacturing process without making too many significant changes to the process. When demands in the manufacturing process change, the reconfigurable workcell needs to be quickly adaptable to cope with these changes.

In our work we developed several hardware components that facilitate reconfiguration and adaptation to new production demands. To comply with the need of SMEs to keep the costs of automation low, we introduce the paradigm of passive reconfigurable components. This paradigm shows to be a more affordable option to high cost off-the-shelf active solutions. 

\subsection{Reconfigurable frame}

The frame of the workcell connects the robot with peripheral modules, which provide different functionalities. The main design requirement for the frame is its stiffness. Stiffness is important for robotic applications because even small frame deformations can result in large positioning errors. On the other hand, the structure must be easily adaptable in order to meet new product demands. With other words, the frame has to be sufficiently flexible so that changes can be made quickly. As the workcell is intended for SMEs, the affordability of such a solution should also be taken into account. 

To fulfil all these requirements, a workcell frame made of rectangular steel beams was chosen. The beams are connected using an innovative system developed for the aeronautical industry called BoxJoint \cite{millar_reconfigurable_2009}. The resulting frame structure is very stiff and comparable with welded joints, while at the same time simple to assemble and modify.

\begin{figure}[t]
    \centering
    \includegraphics[width=.80\linewidth]{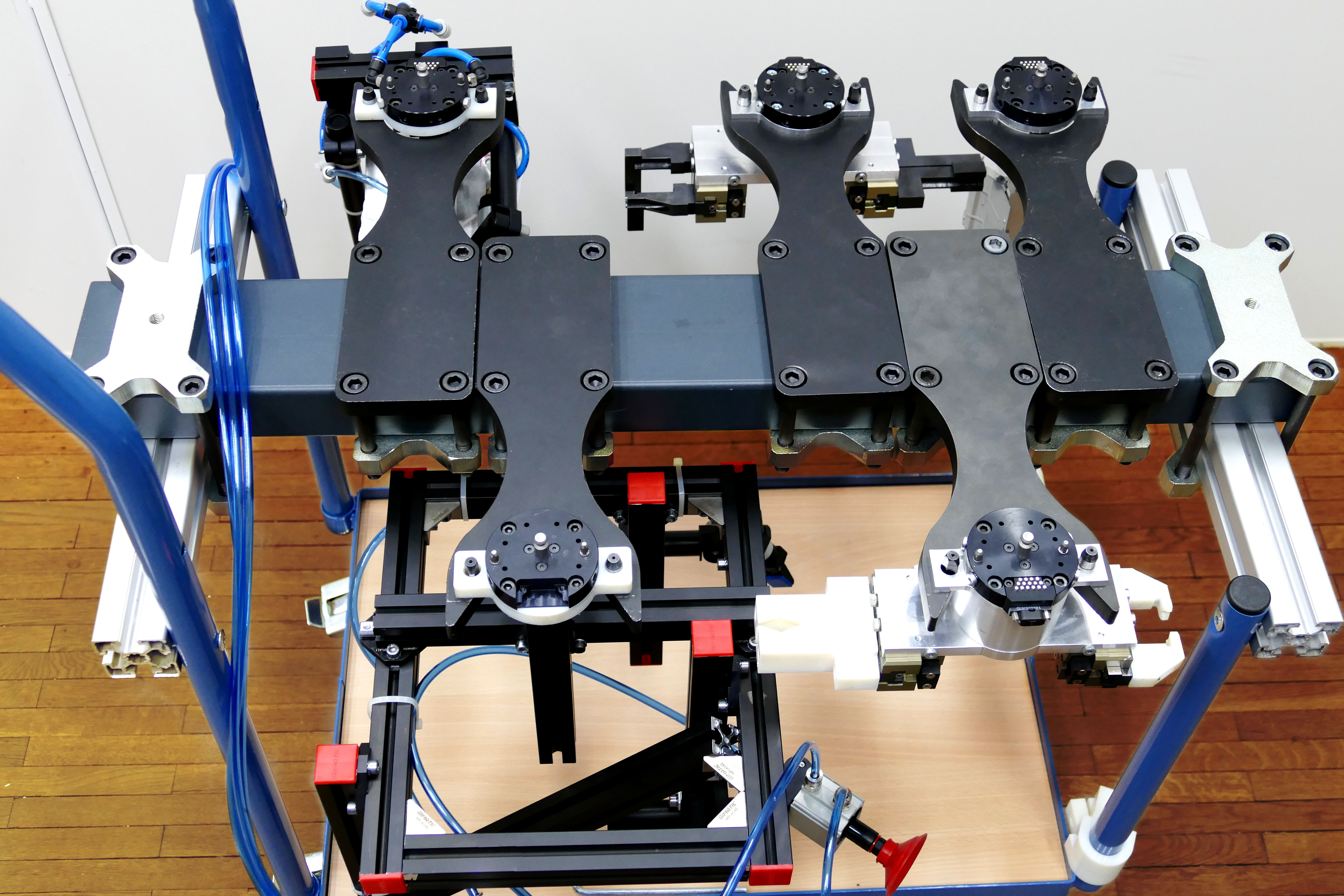}
    \caption{An array of tools available for the robots to equip based on the current step of the assembly. The tools can be attached to the robot using the tool exchange system.}
    \label{fig:tool-rack}
\end{figure}

\subsection{Reconfigurable robot tools}

Traditionally, robot end effectors are equipped with tools specific for a predefined assembly step and are rarely exchanged. While some specialised grippers allow for grasping of objects of different shapes, these variations are usually limited to basic shapes (cubes, spheres, tubes, etc.). To give the robots the needed flexibility to perform a vast array of assembly operations we mounted a tool exchange system to the end effector. This system allows the robots to un/equip tools, which are needed to carry out the current step of the assembly process. This tool exchange system, apart from ensuring a stiff coupling between the robot and the tool, also connects the tool to the pneumatic system, provides electrical power and Ethernet connectivity.

\subsection{Plug \& Produce modularity}

Peripheral modules are crucial for the operation of the reconfigurable robot workcell as they augment the cell with appropriate functionalities. Typically used modules include fixtures, tool storage, material flow management, and other application specific equipment. To ensure a smooth process flow and short reconfiguration times, we need to be able to introduce these modules into the workcell or swap them for another as quickly and with as little interruption to the manufacturing process as possible. 

With this desires in mind, we developed special ``Plug \& Produce'' (PnP) connectors. These are designed to provide quick mechanical coupling with highly repeatable and stiff positioning of the peripheral modules. In order for these peripheral components to be truly modular, they should be self-sufficient to a certain degree, i.\,e. they should be equipped with the appropriate computing, actuation and other capabilities. As only mechanical coupling is not sufficient, the developed PnP connector also provides electrical power, Ethernet connection for data transfer, and pneumatic lines, which can all be used by the equipment contained within the module. This in turn enables the hardware modules to be completely self-sufficient and ready to provide the desired functionality as soon as they are coupled to the main frame. 

\begin{figure}[H]
    \begin{subfigure}{.32\linewidth}
      \centering
      \includegraphics[width=.95\linewidth,trim={10cm 0 15cm 0},clip]{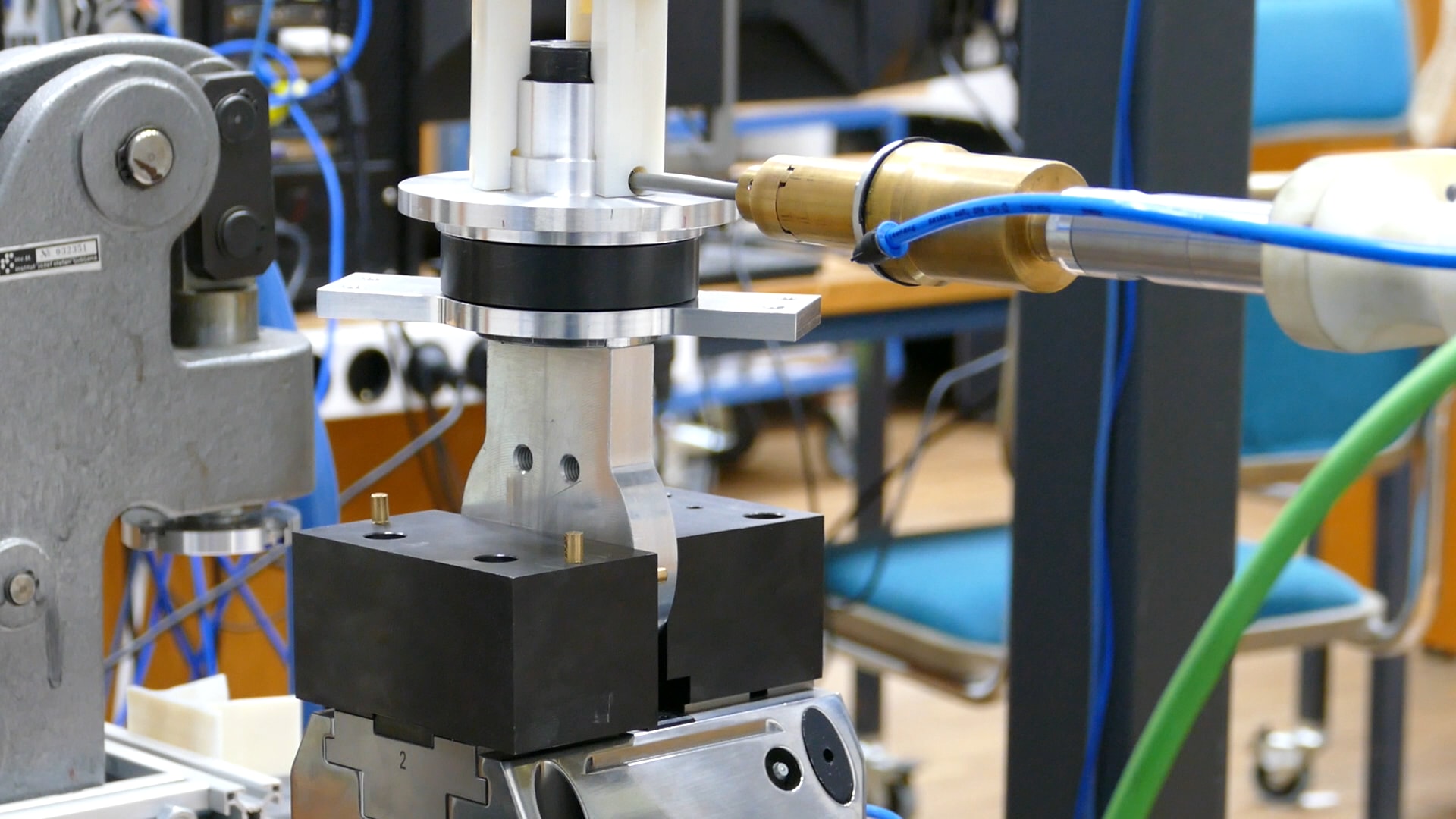}
      \label{fig:rot_2}
    \end{subfigure}
    \begin{subfigure}{.32\linewidth}
      \centering
      \includegraphics[width=.95\linewidth,trim={10cm 0 15cm 0},clip]{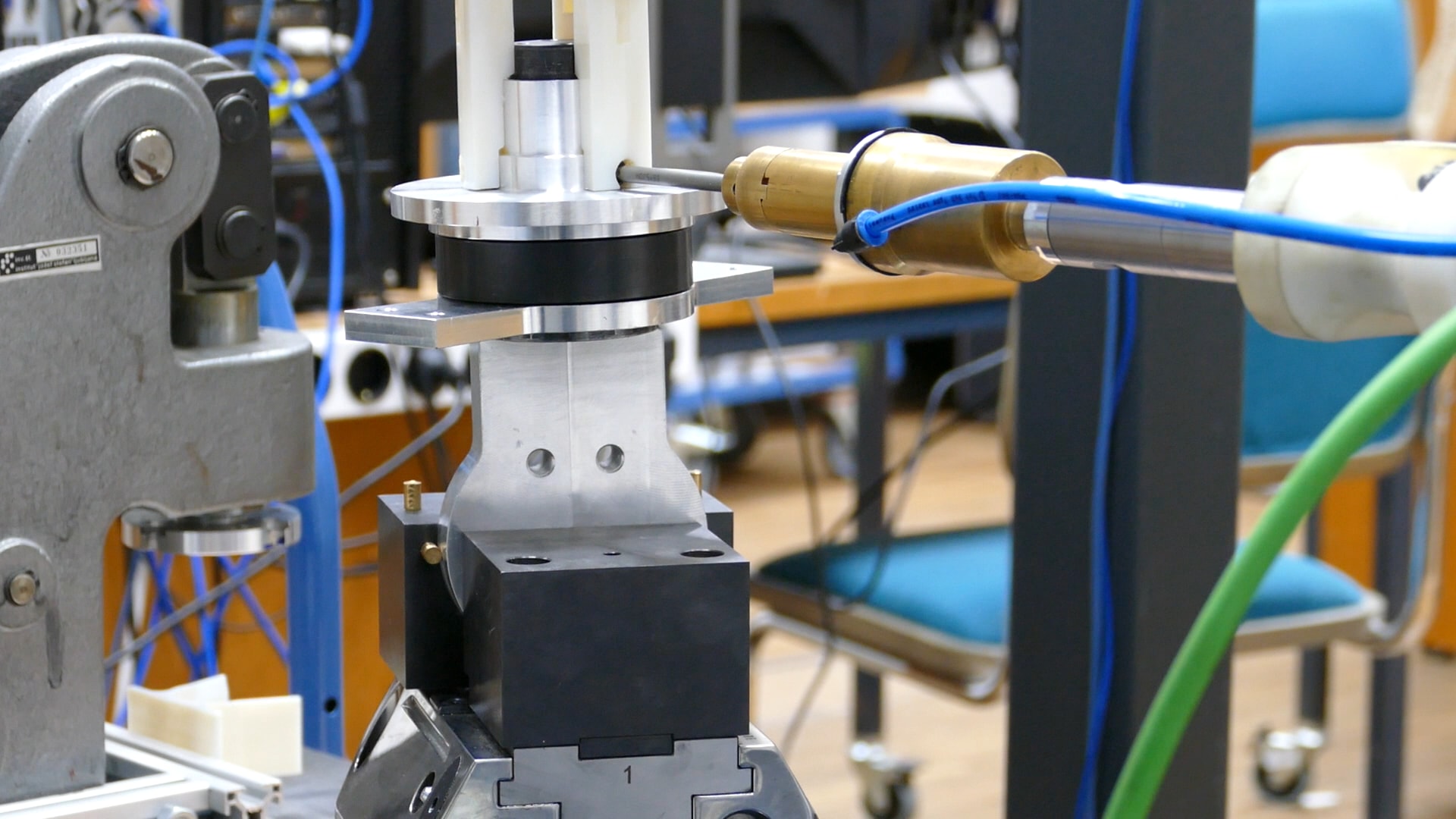}
      \label{fig:rot_3}
    \end{subfigure}
    \begin{subfigure}{.32\linewidth}
      \centering
      \includegraphics[width=.95\linewidth,trim={10cm 0 15cm 0},clip]{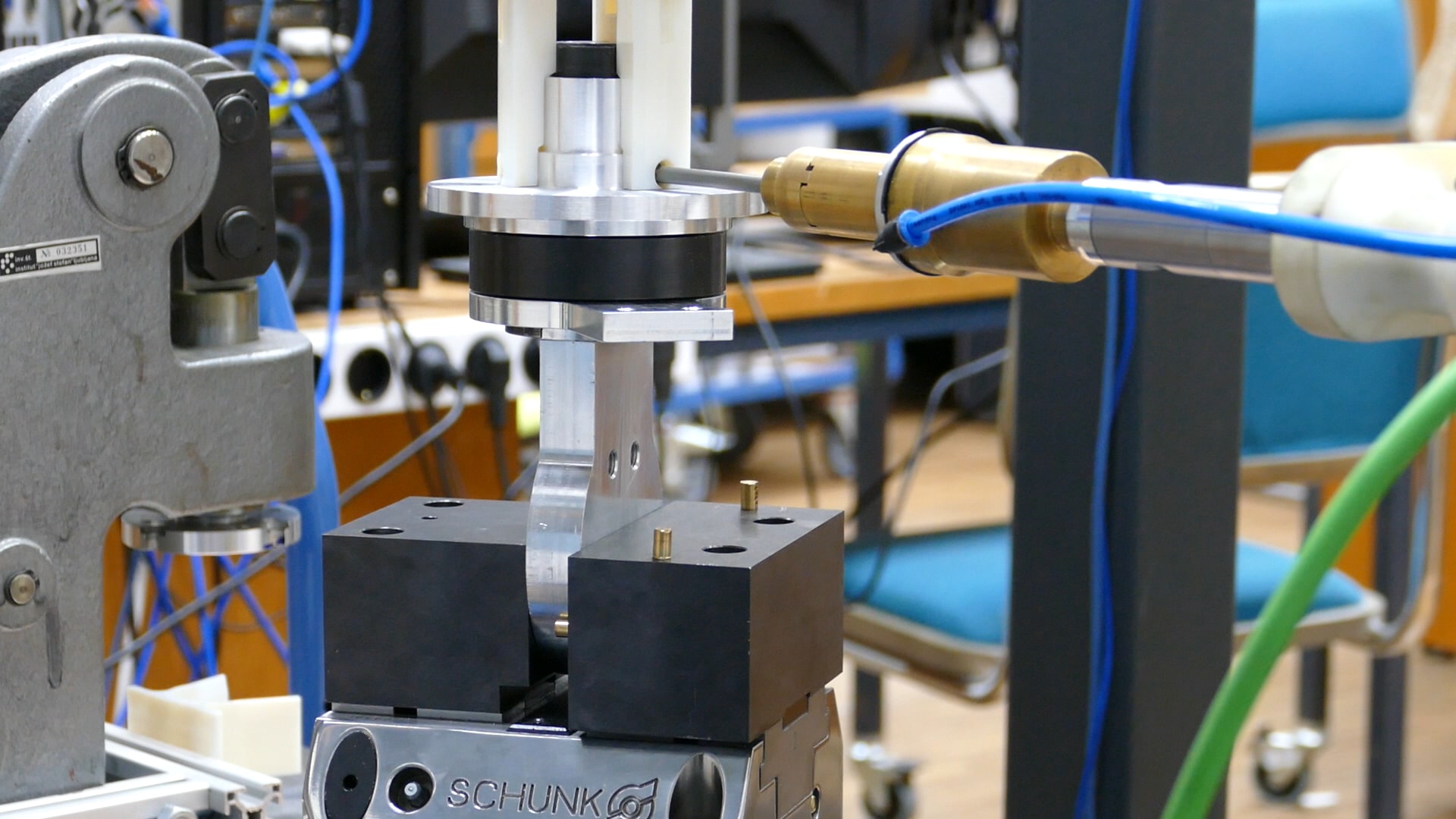}
      \label{fig:rot_4}
    \end{subfigure}
    \centering
    \caption{A sequence (left to right) showcasing the passive rotary table being used to ensure the robot can fasten screws on 3 sides of a workpiece.}
    \label{fig:rot-table}
\end{figure}

\subsection{Passive reconfigurable elements}

While PnP connectors allow us to introduce new modules into the workcell, thus modifying and enriching its functionality, such modules often need to be introduced manually, i.\,e. by a human worker. This type of reconfiguration can therefore not be regarded as fully autonomous. Standard off-the-shelf solutions towards autonomous reconfiguration require additional active components, which can be accompanied by a significantly high price tag. To lower the price of reconfigurable elements, we propose a novel concept of passive reconfigurable hardware components. These passive reconfigurable elements should not contain any actuators or sensing equipment. Instead, since every robot workcell contains a robot, the robot's manipulation and sensing capabilities can be used to carry out reconfiguration and positional sensing.

We evaluated this approach by developing a number of passive hardware components and using them in assembly operations. One such example is a passive rotary table, designed to address the issue of workpiece re-orientation (depicted in Fig. \ref{fig:rot-table}). These rotary table can be, upon releasing the brakes, re-oriented by the robot arm, thus changing orientation of the workpiece placed on top of the table. When the desired new orientation is reached, the brakes are engaged. The robot system stores the last known position of the fixture before fully releasing it.

\begin{figure*}[t]
	\begin{center}
		\includegraphics[width=0.7\textwidth,trim={0.5cm 1.2cm 0.2cm 2.7cm},clip]{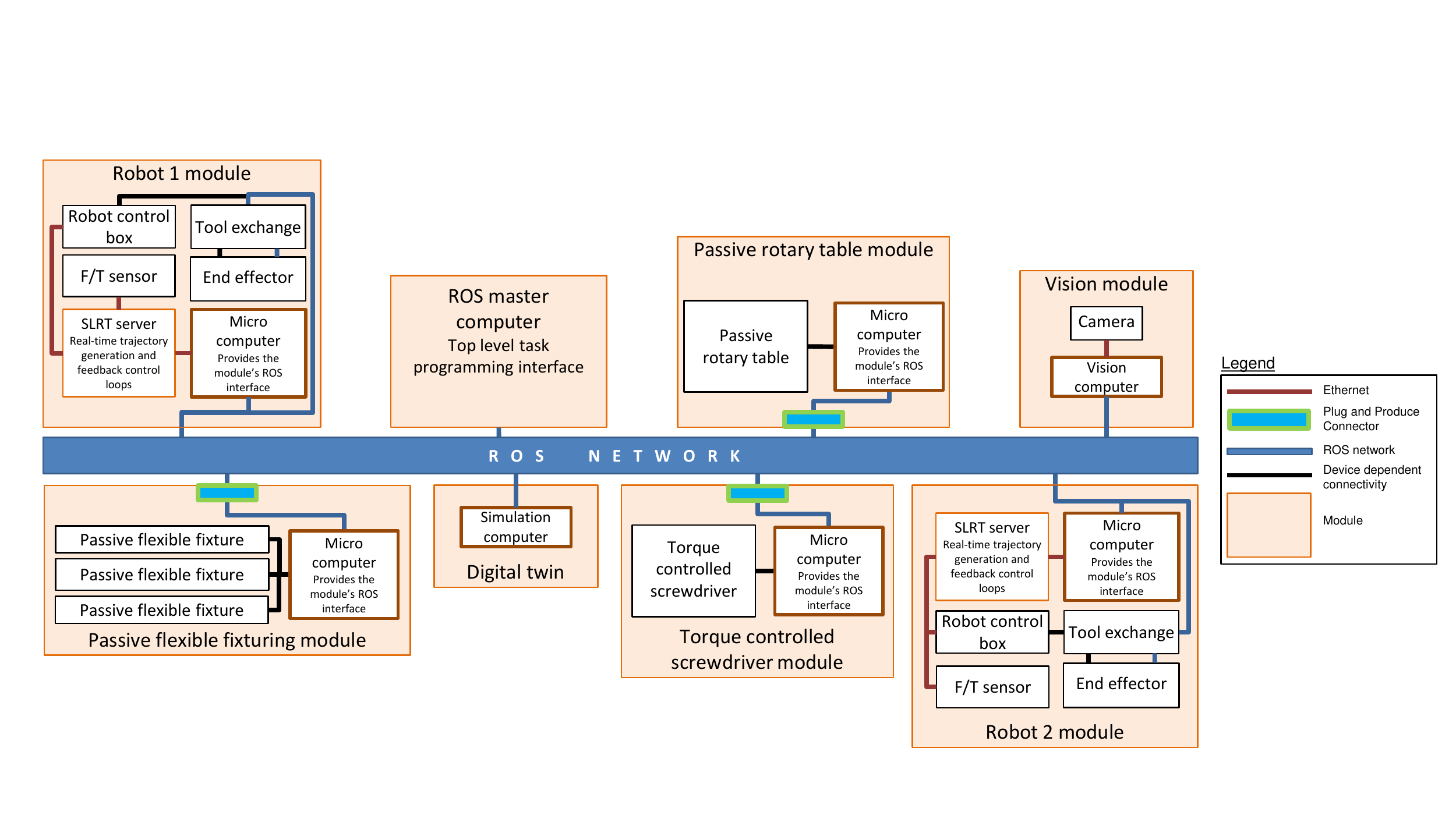}
		\caption{Software architecture for the reconfigurable robot workcell with various software and hardware modules.} \label{fig:software-architecture}
	\end{center}
\end{figure*}


\section{RECONFIGURABLE SOFTWARE}\label{sec:reconf-software}

Besides providing physical connections between peripheral modules, it is also important to ensure connectivity in context of data flow. Each peripheral module should be connected on the same network in order to broadcast its data and receive information and instructions on what action to perform at what time. Furthermore, this data should be parsable by all software components within the system. The software architecture of the proposed cell was is ROS-based, which -- complementary to modular hardware design -- ensures software modularity. An overview of the developed software system architecture of the workcell is depicted in Figure \ref{fig:software-architecture}.

\subsection{Robot workcell ROS backbone}

Simply ensuring the data flow between various modules within the workcell is not enough to adhere to the software modularity requirement of the system. It is necessary to ensure that the data is structured in such a way that it can be parsed by all modules within the system. In this respect, the Robot Operating System (ROS) provides a suitable framework for developing various software components that need to share data over the shared network \cite{quigley_ros_2009}. It essentially eases the integration of different devices into the workcell and allows to control and monitor the workcell as a whole. 

\subsection{ROS-based modules}\label{subsec:ros-modules}

Our driving paradigm is that each module within the workcell should be connected to the ROS network. This introduced a requirement that all modules should be equipped with the computational hardware that is sufficient to run ROS \textit{nodes}, thus exposing each module's data and functionalities to the workcell ROS network (denoted as Micro computer in Fig. \ref{fig:software-architecture}). This enables the modules to be be controlled by the top-level task scheduling software as soon as they are connected to the workcell using the PnP connectors or tool exchange systems described in the previous sections.

\subsection{Low level real-time robot control}

To enable seamless integration with the rest of the hardware components in the workcell, the robots in the developed workcell can also communicate via ROS. Most of the industrial robots are equipped with a control box that provides real-time control functionalities. However, these control boxes do not support running ROS nodes, therefore we need to implement a special communication layer to connect the robot with the rest of the ROS network. We therefore developed an abstraction layer that supports switching between different types of robots. This layer provides a number of trajectory and feedback control strategies independently of the selected robot and enable the programming of new strategies via a suitable control interface. 
This design decision is compliant with the paradigm that the robot is a module within the workcell and should therefore be easily replaceable. In addition, this layer makes the cell non-robot-specific, which add to the overall modularity of the cell.

\section{QUICK SET-UP OF ASSEMBLY SEQUENCES}\label{sec:task-programming}

Compiling the assembly process sequence to the workcell task-level scheduler usually takes a significant portion of the setup time. Shortening of this time is therefore very important in order to allow fast setup and short reconfiguration times, this compilation should be as fast as possible. To tackle this, we developed and implemented a set of technologies in order to facilitate and accelerate the programming of robot workcell assembly operations.

\subsection{Assembly skill programming by demonstration}

Defining the robot motions for a complete assembly process can be difficult and time consuming for non-expert users. Programming by Demonstration (PbD) provides a methodology to define these motions in a natural way rather than coding complex programs in a robot-oriented programming language \cite{dillmann_2004,billard_robot_2008, argall_survey_2009}. In our work, we identified the need to offer at least two approaches to PbD: kinesthetic teaching and remote guidance.

The first approach allows the user to move the robot through its workspace by physically guiding it through the desired motion. Kinesthetic guidance has been widely adopted in collaborative robots as it is the most effective with modern torque-controlled robots \cite{hersch_dynamical_2008}. While kinesthetic teaching excels in intuitiveness, the quality of the dynamic model has a considerable impact on the physical effort needed in order to move the robot. Consequently, the more effort is needed to manually guide the robot, the harder it is to guide it along a smooth path and position it precisely in a desired configuration.

These drawbacks of kinesthetic guidance represent an inconvenience when working towards methods to shorten times of robot programming. A badly implemented kinesthetic guidance can result in the user spending a significant amount of time trying to achieve the desired robot movement or configuration. In response to these drawbacks, we developed a remote control interface that allows the user to control the robot by interacting with a consumer grade joystick. To achieve smooth control of the robot, we mapped the displacement of the analogue sticks to the Cartesian space velocities.

\subsection{Database of assembly skills}

When developing the programming by demonstration framework, we wanted the acquired assembly skills to be accessible throughout the entire software framework of the workcell. For this reason, we integrated the MongoDB database into our system, where the taught skills are stored. Acquiring a new skill therefore means making a new named entry into the MongoDB database. It is then possible to define an assembly sequence that reads the named database entries from the database and moves the robots accordingly. One of the main benefits of having these skills saved as named entries is that it allows for quick reconfiguration in terms of changing certain skills. It is sufficient to overwrite the entry with a modified skill to update it without changing the top-level assembly sequence program.

\subsection{State machine assembler}

To further accelerate the programming process of the workcell assembly sequence, an engine for state machine code generation was developed. There are numerous ROS-based packages aimed at facilitating the high-level task programming by using state machines. However, defining complex robot behaviours with these tools requires a programmer to dedicate his attention, not only to the structure of the state machine, but also to the boilerplate code and programming language syntax. To expedite and enhance this process, a method for code generation, templating and meta-scripting was developed and is presented in detail in our previous work \cite{ridge_rapid_2017}.

\section{CONCLUSIONS AND VALIDATION}\label{sec:conclusion}

In this project we developed a highly reconfigurable robot workcell with innovative hardware concepts and components with a ROS-based software backbone. In our work we strove not only to provide means for autonomous reconfiguration to adapt to production changes, but also to shorten set-up times by implementing various programming by demonstration technologies. The developed proposed paradigms, the underlying technology and the overall robot workcell has been extensively evaluated through implementing five use-cases from different fields of industry in the relevant environment. These use-cases range from the (1) assembly of automotive headlights, (2) the assembly of linear drives, (3) the assembly of a robotic gripper, (4) assembly of airport runway lights and finally the (5) assembly of printed circuit boards (PCBs).
By doing so, we were able to prove the overall performance of the developed cell and acquire the first reference key performance indicators. Some of the key equipment stayed the same (i.e. robots, tool rack, etc.), however other parts of the cell were reconfigured according to the requirements of each experiment. Some application-specific periphery modules were either added or removed.

\section*{Acknowledgment}
This work has been funded by the Horizon 2020 ICT-FoF Innovation Action no 680431,  ReconCell (A Reconfigurable robot workCell for fast set-up of automated assembly processes in SMEs).

\addtolength{\textheight}{-12cm}   

\bibliographystyle{ieeetr}
\bibliography{bibliography}

\end{document}